\documentclass[sigconf, nonacm]{acmart}

\AtBeginDocument{%
  }

\usepackage{amsmath} 
\usepackage{mathtools} 
\usepackage{amsthm} 

\usepackage{subcaption} 
\usepackage{arydshln} 
\usepackage[table]{xcolor} 
\usepackage{enumitem} 

\usepackage{booktabs} 
\usepackage{multirow} 
\usepackage{tabularray} 
\usepackage{nicematrix} 
\usepackage{graphicx} 

\usepackage{tabularx} 

\begin{document}

\title{Scaling Agentic Capabilities via Grounded Interaction Synthesis}

\author{Wenhang Shi}
\affiliation{%
  \institution{Renmin University of China}
  \city{Beijing}
  \country{China}
}
\email{wenhangshi@ruc.edu.cn}

\author{Jinhao Dong}
\authornote{Corresponding author}
\affiliation{%
  \institution{Renmin University of China}
  \city{Beijing}
  \country{China}
}
\email{dongjinhao@ruc.edu.cn}

\author{Yiren Chen}
\affiliation{%
  \institution{Peking University}
  \city{Beijing}
  \country{China}
}
\email{yrchen92@pku.edu.cn}

\author{Zhe Zhao}
\affiliation{%
  \institution{Tencent}
  \city{Beijing}
  \country{China}
}
\email{yrchen92@pku.edu.cn}

\author{Shuqing Bian}
\affiliation{%
  \institution{Tencent}
  \city{Beijing}
  \country{China}
}
\email{shuqingbian@gmail.com}

\author{Wei Lu}
\affiliation{%
  \institution{Renmin University of China}
  \city{Beijing}
  \country{China}
}
\email{lu-wei@ruc.edu.cn}

\author{Xiaoyong Du}
\affiliation{%
  \institution{Renmin University of China}
  \city{Beijing}
  \country{China}
}
\email{duyong@ruc.edu.cn}



\begin{abstract}
General agentic intelligence hinges on the ability to interact with diverse real-world tools to complete complex tasks, a capability fundamentally tied to the quality of interaction data. 
To bypass the prohibitive costs of human annotation, prevailing paradigms depend entirely on Large Language Models (LLMs) to scale the synthesis of agentic environments and tasks.
However, such unconstrained generation often degenerates into biased random sampling of LLMs' internal priors, failing to capture the diversity and difficulty of real-world domains or construct high-fidelity, long-horizon tasks.
In this work, we introduce Grounded Agentic Interaction Synthesis (GAIS), a framework that automates the scalable construction of diverse environments and complex tasks via a two-phase grounding mechanism.
Specifically, we construct protocol-anchored environments derived from real-world Model Context Protocol (MCP) servers to ensure functional diversity and difficulty. Subsequently, we employ structure-guided planning to navigate these environments, actively enforcing logical dependencies and adversarial policies to  generate complex tasks.
Experiments on BFCL, $\tau^2$-Bench, and ACEBench demonstrate that GAIS-synthesized data significantly outperforms state-of-the-art baselines, enabling base models to match or even surpass their official instruction-tuned counterparts.
Furthermore, GAIS exhibits superior data efficiency and scalability, achieving exceptional capabilities with significantly less data while maintaining continuous growth where baselines stagnate. Our code and data are publicly available at \href{https://github.com/Eric8932/GAIS}{https://github.com/Eric8932/GAIS}.

\end{abstract}



\maketitle

\section{Introduction}

General agentic intelligence hinges on the ability to interact with diverse real-world tools to accomplish complex tasks \cite{DBLP:journals/corr/abs-2406-12045,barres2025tau2,DBLP:conf/icml/PatilMYJSSG25}.
To foster these capabilities in Large Language Models (LLMs), training paradigms are shifting from static corpora to learning from interactive experience \cite{silver2025welcome,DBLP:conf/iclr/QinLYZYLLCTQZHT24}.
Constructing high-quality interaction data necessitates two essential components: executable tool environments, which define the action space, and task specifications anchored to specific tool chains, which shape the user-assistant trajectories \cite{DBLP:journals/csur/QinHLCDCZZHXHFSWQTZLSXZ25}.
Ideally, both components should be harvested from authentic human interactions with real-world applications to ensure fidelity. However, the prohibitive costs of deploying authentic environments and curating high-fidelity tasks severely limit the scalability of the manual acquisition pipeline. 



To scale the synthesis of interaction data, previous research predominantly rely on LLMs. For environment construction, they often utilize LLMs to synthesize the entire pipeline, sequentially generating domains, applications, toolsets, and even simulating executable implementations \cite{DBLP:journals/corr/abs-2509-13311,basant2025nvidia}.
While some approaches incorporate open APIs to ensure realism, they are typically confined to simple, read-only retrieval tools \cite{DBLP:conf/iclr/QinLYZYLLCTQZHT24,DBLP:journals/corr/abs-2306-06624,DBLP:journals/corr/abs-2507-16044}.
Regarding task generation, existing methods typically prompt LLMs to directly sample tool chains and corresponding user queries from the available toolset \cite{DBLP:conf/iclr/Liu0ZHYL0GLY0WN25,team2025kimi,team2025longcat}.
While such automation enables massive scaling, this unconstrained synthesis fundamentally degenerates into biased sampling of the LLMs' inherent distributions.
Through a systematic analysis of existing datasets, we reveal that without structural grounding, LLMs tend to generate semantically repetitive tools and tasks characterized by shallow interaction chains.
Specifically, over 90\% of samples consist of simple information retrieval tool environments paired with low-complexity tasks. Consequently, these approaches fail to simulate the diverse, complex, and long-horizon interactions required to foster robust agentic capabilities.

To bridge the gap between scalability and data quality, we introduce Grounded Agentic Interaction Synthesis (GAIS), a framework that automates the scalable construction of diverse environments and complex tasks via a two-phase grounding mechanism.
First, we construct protocol-anchored environments by mining the real Model Context Protocol (MCP) ecosystem \cite{anthropic2024mcp,hou2025model}.
To circumvent the prohibitive overhead of MCP tool deployment, we leverage LLMs to transform authentic tool implementations into  executable Python code. 
This process is designed to retain high fidelity to the real-world application logic, resulting in environments that exhibit high functional diversity and difficulty.
Second, by grounding task construction in complex tool dependencies, we generate cognitively challenging scenarios.
Moving beyond random environment exploration, we explicitly plan execution paths around complex write operations, necessitating intricate inter-tool dependencies.
Furthermore, we inject adversarial constraints that require handling restrictions or refusing requests.
Based on the grounded environments and tasks, we simulate human-assistant interactions to generate long-horizon trajectories that mirror the friction and complexity of real-world workflows.



Extensive experiments on agentic benchmarks such as BFCL \cite{DBLP:conf/icml/PatilMYJSSG25}, $\tau^2$-Bench \cite{barres2025tau2}, and ACEBench \cite{chen2025acebench} demonstrate the efficacy of our GAIS-synthesized data. 
Across various scales of Qwen3-Base series, models trained on our data significantly outperform state-of-the-art baselines, notably enabling base models to match or even surpass the performance of their official instruction-tuned counterparts.
Furthermore, our approach exhibits superior data efficiency and scalability, achieving exceptional capabilities with significantly less data while sustaining continuous performance gains where existing methods plateau.
Crucially, ablation studies confirm that environment diversity and task complexity are essential for these gains, validating the necessity of grounded synthesis for scaling agentic intelligence.
Our contributions are summarized as follows:
\begin{itemize}[topsep=-1pt, itemsep=-0.5pt, leftmargin=8.5pt]
\item \textbf{Systematic analysis of LLM-synthesis bias.} We provide a systematic analysis of existing tool-learning datasets synthesized predominantly via LLMs, revealing critical deficiencies in environment diversity and difficulty, alongside a prevalence of trivial, short-horizon tasks.
\item \textbf{Grounded data synthesis framework.} We propose a novel framework that grounds data generation in protocol-anchored environments and structure-guided tasks, automating the creation of high-fidelity and complex human-assistant interactions.
\item \textbf{Superior data effectiveness and efficiency.} We demonstrate that models trained on our synthesized data outperform state-of-the-art baselines, achieving superior capabilities with less data while sustaining more continuous performance growth.

\end{itemize}

\section{Related Work}

\subsection{Tool-Use Environments}

Existing approaches for constructing tool-use environments generally consist of two components: the tool source and the implementation mechanism.
Regarding the source, real-world applications accessed via APIs or the MCP servers offer the gold environment for fidelity \cite{DBLP:conf/iclr/QinLYZYLLCTQZHT24,DBLP:journals/corr/abs-2306-06624,DBLP:journals/corr/abs-2507-16044}. However, readily accessible APIs are often restricted to simple read-only tools, while complex services incur prohibitive overheads related to deployment, authentication, and stability. 
Conversely, LLM-synthesized approaches circumvent these costs by sequentially generating domains, applications and toolsets \cite{team2025kimi,basant2025nvidia,team2025longcat}. Yet, this unconstrained generation often yields homogeneous environments lacking in functional difficulty.
Furthermore, in the absence of live endpoints, the tool implementation becomes the second bottleneck. Some methods rely on LLM-based simulators to mock feedback \cite{DBLP:conf/naacl/LuHZANBMMLYWP25,basant2025nvidia,team2025longcat}, a scalable but unreliable approach prone to hallucination and inconsistent state tracking. Others construct offline code environments \cite{DBLP:journals/corr/abs-2509-13311,DBLP:journals/corr/abs-2406-12045,DBLP:journals/corr/abs-2508-08791}, which ensure deterministic dynamics but typically demand extensive manual engineering.
To reconcile scalability with fidelity, our framework automatically converts rich real-world MCP servers into executable Python functions via a test-driven iterative pipeline, securing both functional diversity and deterministic reliability.

\subsection{Tool-Learning Tasks}
Generating agentic tasks that entail specific tool-call chains is also critical for gathering high-quality user-assistant interactions.
Existing methods typically leverage LLMs to generate user intents based on sampled tool subsets \cite{DBLP:journals/csur/QinHLCDCZZHXHFSWQTZLSXZ25}, which can be categorized by their sampling strategies.
The most basic approach relies on direct random sampling from the toolset \cite{DBLP:conf/iclr/QinLYZYLLCTQZHT24,DBLP:conf/iclr/Liu0ZHYL0GLY0WN25,team2025kimi,basant2025nvidia}. Due to the disregard of tool dependencies, it typically yields trivial tasks with shallow interactions that fail to reflect the complexity of real-world workflows. 
While some works structure toolsets into dependency graphs \cite{DBLP:journals/corr/abs-2509-13311,team2025longcat}, their random walks tend to gravitate towards dominant simple nodes, similarly resulting in limited task complexity.
Recent improvements anchor these random walks around write tools to induce more complex state changes \cite{DBLP:journals/corr/abs-2504-03601,DBLP:conf/acl/Yin0HYJCGLCLPP25}. However, the effectiveness of these methods is constrained by the underutilization of complex inter-tool dependencies.
In contrast, our GAIS framework explicitly plans execution paths around deep dependencies and injects adversarial policies that mandate constraint handling, ensuring the generation of high-fidelity complex tasks.


\section{Quantitative Analysis of Agentic Datasets}\label{sec_quantitative_analysis}
The quality of user-assistant interaction data is fundamentally dependent on the functional diversity and difficulty of the underlying tool environments, alongside the complexity of the constituent agentic tasks.
In this section, we quantitatively analyze and compare these dimensions across various leading open-source datasets and our GAIS-synthesized data, revealing the structural biases inherent in unconstrained LLM-based synthesis.
We select ToolLLM \cite{DBLP:conf/iclr/QinLYZYLLCTQZHT24} as the representative of tool-learning data based on real-world APIs. To represent the paradigm of purely LLM-synthesized data, we examine ToolACE \cite{DBLP:conf/iclr/Liu0ZHYL0GLY0WN25} and Nemotron (the tool-calling subset of the Nemotron-Post-Training-Dataset-v1) \cite{basant2025nvidia}.
For scalable categorization and scoring, we employ Gemini-2.5-Pro~\cite{comanici2025gemini} as an automated annotator. Validating against 200 expert-annotated samples (50 per dataset), Gemini achieves 97\%, 98\%, 100\%, and 98\% agreement with at least one of two experts across tool type, difficulty, dependency, and task complexity, respectively, confirming the reliability.


\begin{figure}[t]
\includegraphics[width=\columnwidth]{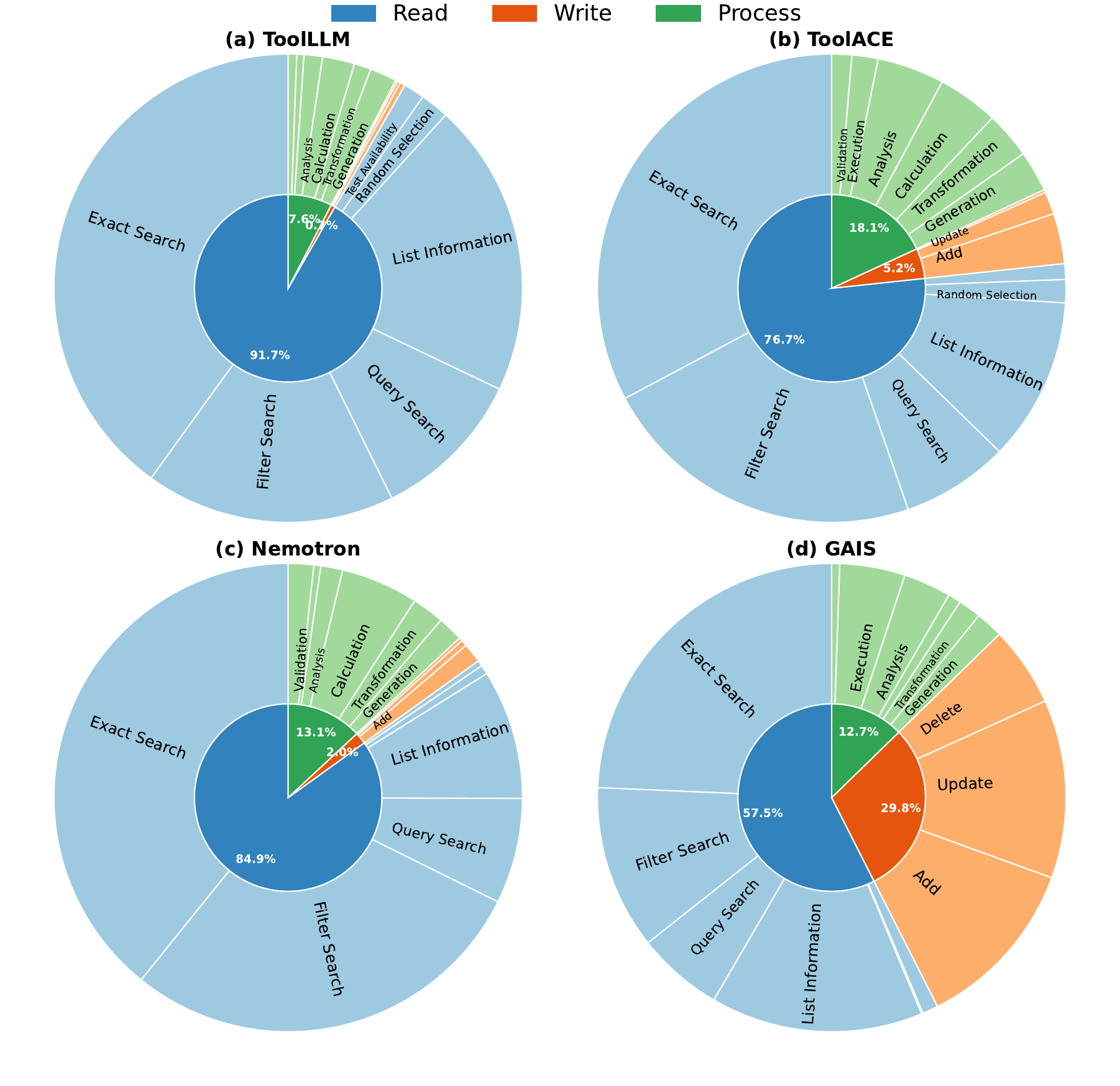}
\Description[Sunburst charts comparing tool type distributions across ToolLLM, ToolACE, Nemotron and GAIS.]{
The figure presents four sunburst charts (a, b, c, d) illustrating the distribution of tool types categorized into "Read" (inner ring, dominant), "Write", and "Process", along with their specific sub-categories in the outer ring.

Charts (a) ToolLLM, (b) ToolACE and (c) Nemotron, exhibit a heavily skewed distribution dominated by "Read" tools. Specifically:
- In (a) ToolLLM, "Read" tools constitute 76.7\% of the total.
- In (b) ToolACE, "Read" tools make up 84.9\%.
- In (c) Nemotron, "Read" tools account for 91.7\%.
The "Write" and "Process" categories in these datasets are visibly minor segments.

In contrast, Chart (d) GAIS shows a significantly more balanced distribution. The "Read" category is reduced to 57.5\%, while "Write" and "Process" tools encompass substantial portions of 29.8\% and 12.7\% respectively. The outer ring of GAIS details diverse operations such as "Update", "Add", and "Delete" under the Write category, and "Analysis", "Execution", and "Transformation" under the Process category, highlighting a richer functional diversity compared to the baselines.
}
\caption{Distribution of tool types within environments derived from real-world APIs (a), fully LLM synthesis (b, c), and our protocol-anchored construction (d).}
\vspace{-0.3cm}
\label{fig_diversity}
\end{figure}

\subsection{Environment Diversity}\label{subsec_quantify_diversity}
The functional diversity of tools within an environment essentially dictates the breadth of feasible tasks and the depth of resulting user-assistant interactions.
To quantify environment diversity across datasets, we deduplicate tool definitions within the training samples and categorize each tool into three distinct types: read, write, and process. 
Read tools retrieve information from an existing data source (e.g., database queries, weather fetching). Write tools modify system states (e.g., creating records, deleting files, updating posts). Process tools perform computational transformations to generate new outputs (e.g., image resizing, mathematical computation). A detailed taxonomy and classification criteria are in Appendix~\ref{appen:subsec_tool_taxonomy}.


The statistical distribution for each dataset is presented in Figure \ref{fig_diversity}.
As observed, read tools dominate existing datasets, constituting over 75\% and in some cases exceeding 90\% of the total tool distribution.
This indicates that whether sourced from public APIs or synthesized entirely by LLMs, current environments are heavily skewed towards simple retrieval tools.
We attribute this imbalance to distinct factors in each paradigm.
For real-world APIs, while complex write and process tools exist, they typically entail prohibitive authentication, deployment, or registration, leaving only simple read APIs readily accessible.
In fully LLM-synthesized environments, unconstrained generation fundamentally samples the LLMs' prior distribution, which naturally gravitates towards common, low-entropy retrieval functions.
In contrast, our protocol-anchored environments exhibits a more balanced distribution across all three categories, encompassing not only common retrieval tools but also a rich set of state-modifying and data-processing functions. 
This demonstrates that while LLM synthesis is effective for scaling environments, explicit structural grounding is indispensable to break the retrieval-centric bias inherent in LLM-based generation.

\subsection{Environment Difficulty}\label{subsec_quantify_difficulty}
Beyond semantic diversity, the interactive difficulty of an environment, defined by the operational difficulty of its constituent tools, establishes the upper bound for task complexity.
To quantify this, we employ LLMs to score tools on a 1-to-5 difficulty scale \cite{comanici2025gemini}. The scoring primarily evaluates parameter complexity, domain knowledge requirements, and the intricacy of execution outcomes, ranging from trivial, parameter-light abstractions (scores 1-2) to complex, state-altering operations requiring precise control (scores 4-5). Detailed scoring guidelines are in Appendix Table \ref{app:tab_difficulty_scoring}.

\begin{figure}[t]
\centering
\includegraphics[width=\columnwidth]{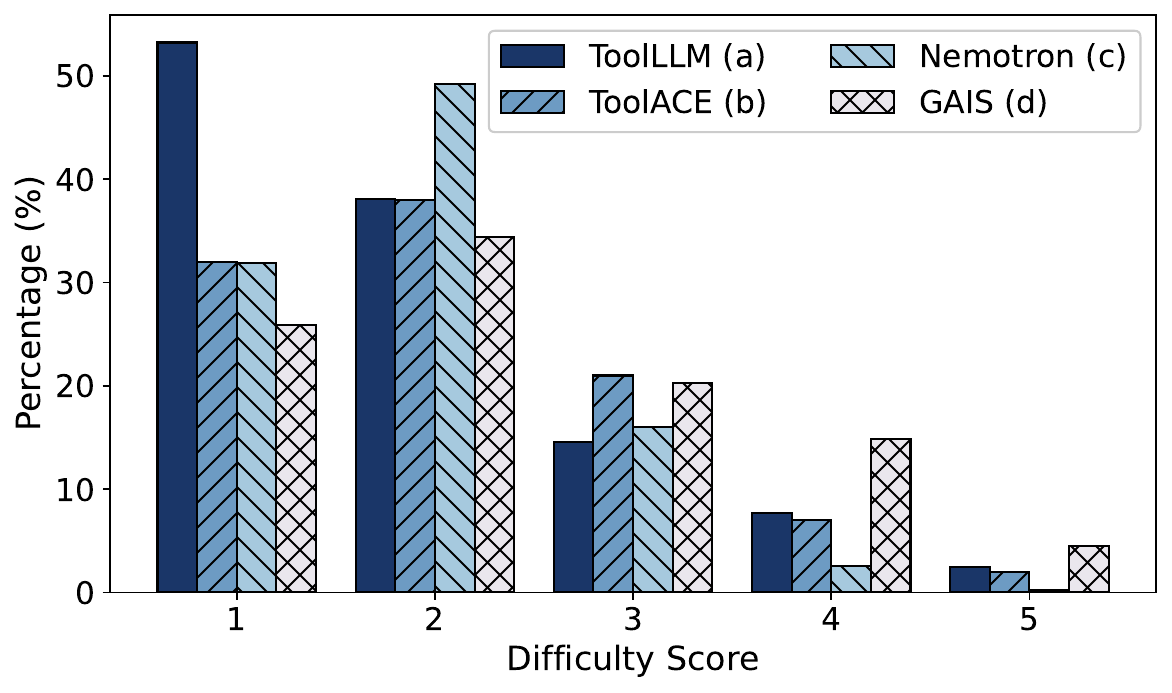}
\vspace{-0.3cm}
\Description[Bar chart comparing the tool difficulty distribution across ToolLLM, Nemotron, ToolACE, and GAIS.]{
The figure presents a bar chart illustrating the distribution of tool difficulty scores (ranging from 1 to 5) across four datasets: ToolLLM, Nemotron, ToolACE, and GAIS. The y-axis represents the percentage of tools, scaled from 0\% to 50\%.

The three baseline datasets (ToolLLM, Nemotron, and ToolACE) exhibit a heavily skewed distribution towards low difficulty. Tools with difficulty scores of 1 and 2 dominate these environments, collectively accounting for over 80\% of the total volume. The proportion of tools drops rapidly as difficulty increases, with scores 4 and 5 being negligible.

In contrast, the GAIS dataset (Ours) shows a significantly more balanced distribution. It maintains a moderate representation of simple tools (scores 1-2) while featuring a substantial proportion of high-complexity tools (scores 4-5). This indicates that GAIS environments encompass a wider range of functional depth, from simple utilities to complex, state-altering operations, unlike the predominantly trivial tools found in the baselines.
}
\caption{Distribution of tool difficulty score within environments derived from real-world APIs (a), fully LLM synthesis (b, c), and our protocol-anchored construction (d).}
\label{fig_difficulty}
\end{figure}

The distribution of tool difficulty is illustrated in Figure \ref{fig_difficulty}. In existing datasets, low-difficulty tools (scores 1-2) dominate, constituting over 80\% of the total volume, with the frequency of tools sharply declining as difficulty increases.
The prevalent high-level function abstractions require minimal cognitive effort to utilize, often lacking the strict formatting constraints, parameter dependencies, and specialized domain logic characteristic of professional applications.
Consequently, such shallow environments fundamentally constrain the generation of trajectories simulating the friction and error-proneness of real-world interactions, where agents must navigate complex, high-stakes tool executions.
In contrast, our protocol-anchored environments exhibit a significantly more balanced distribution, encompassing not only commonplace utilities but also a substantial proportion of high-difficulty tools.
This difficulty distribution aligns closely with the diversity analysis. Since read tools are inherently less complex than write or process tools, the overwhelm of read tools in existing datasets naturally results in a scarcity of functional difficulty.
Given that modern LLMs are already proficient at synthesizing simple retrieval tools, the critical value of tool-learning data lies in the inclusion of high-difficulty operations, highlighting  the necessity of introducing external structural grounding to scale agentic data generation.


\subsection{Task Complexity}\label{subsec_quantify_task}
\begin{figure}[t]
\centering
\includegraphics[width=\columnwidth]{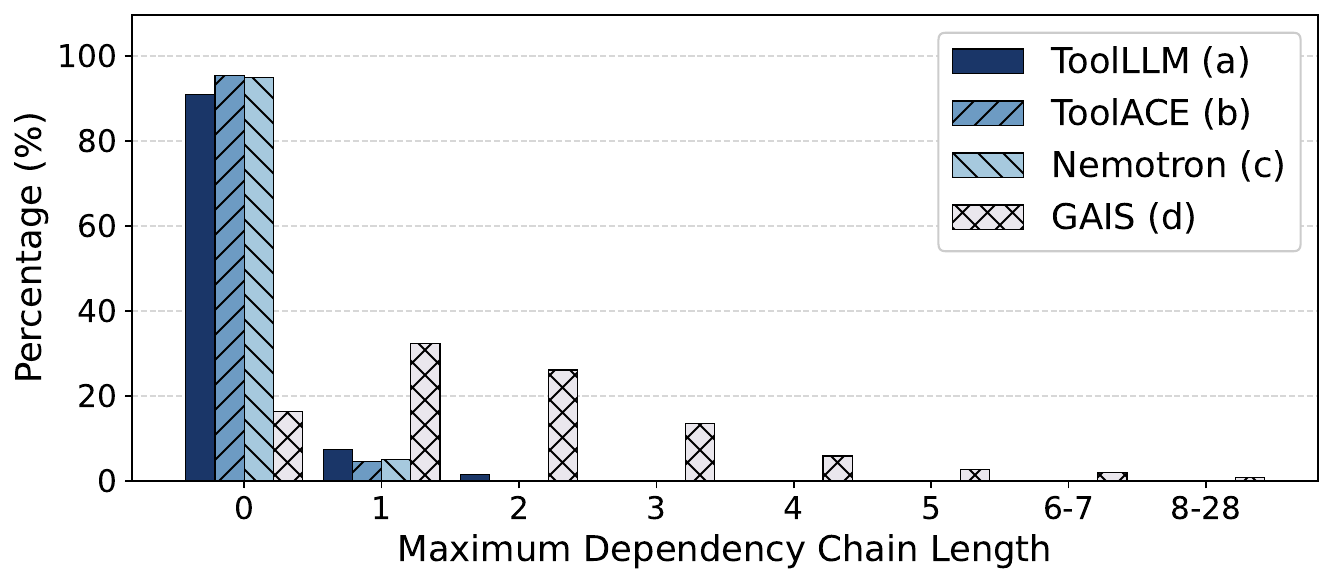}
\Description[Bar chart comparing the maximum dependency chain length across ToolLLM, Nemotron, ToolACE, and GAIS.]{
The figure displays a grouped bar chart illustrating the distribution of maximum dependency chain lengths (x-axis categories: 0, 1, 2, 3, 4, 5, 6-7, 8-28) across four datasets. The y-axis represents the percentage of trajectories, ranging from 0\% to 100\%.

The three baseline datasets—ToolLLM (a), Nemotron (b), and ToolACE (c)—show a heavily skewed distribution dominated by zero-dependency tasks. Specifically:
- Nemotron and ToolACE are nearly identical, with approximately 90\% of their trajectories having a dependency chain length of 0.
- ToolLLM also peaks at length 0 (approx. 65\%) and drops rapidly, with negligible representation beyond length 3.

In contrast, the GAIS dataset (d) exhibits a significantly broader and flatter distribution. The percentage of length-0 tasks is low (below 20\%). The distribution shows substantial volume in deeper reasoning chains, peaking around lengths 3 and 4 (each constituting roughly 15-20\%). Notably, GAIS is the only dataset with visible bars in the long-tail categories of 6-7 and 8-28, indicating its capability to support complex, multi-step tasks.
}
\caption{Distribution of the \textbf{maximum dependency chain length} within user-assistant trajectories. Comparisons cover fully LLM-synthesized pipelines (a, b, c) and our structure-guided task generation (d).}
\label{fig_depen_traj}
\end{figure}

The quality of user-assistant interaction trajectories depends not only on the diversity and difficulty of the underlying environment but crucially on the complexity of the agentic tasks synthesized within it.
To quantify task complexity, we employ a comprehensive suite of metrics to analyze the complexity of the resulting interaction samples.
Beyond standard descriptive statistics, such as the average number of interaction turns and tool invocations, we focus specifically on the structural complexity of the tool dependency chains, which reflects the sequential reasoning and causal logic required for real-world problem-solving.
To extract these chains, we prompt LLMs to identify pairwise causal relationships between tool invocations, capturing dependencies through both data flow (explicit parameter passing) and control flow (logical semantic premises).
By modeling the tool execution sequence of each training sample as a directed acyclic graph, we calculate the maximum dependency chain length and total dependency count to serve as quantitative complexity proxies for reasoning depth and logical density, respectively.



Figure \ref{fig_depen_traj} illustrates the distribution of the maximum dependency chain length, revealing a stark contrast between datasets.
In existing datasets, over 90\% of samples feature tool calls with zero or minimal dependencies (lengths of 0 to 2). This indicates that random environment exploration fails to leverage the latent causal structures between tools, resulting in tasks that appear as loose collections of independent actions rather than cohesive workflows.
In contrast, the trajectories in our dataset not only exhibits a significantly higher prevalence of tool dependencies but also heavily populates the long tail of deep reasoning chains.
Furthermore, as shown in Table \ref{tab_depen_traj}, our method achieves superior performance across other complexity metrics, including average interaction turns, tool invocation count, and dependency count. 
These results validate the importance of our structure-guided task generation: without structural guidance, LLMs gravitate towards generating fragmented and simplistic tasks even when capable of more complex reasoning.


To demonstrate that the superior complexity of our interaction samples is driven primarily by our task generation strategy rather than the environments, we further analyze the complexity of the underlying environments in Appendix Table \ref{app:tab_depen_env} and Figure \ref{app:fig_depen_env}, specifically examining the distribution of available tool count, dependency count, and maximum chain length. 
While our environments show the highest complexity, the API-based ToolLLM environments also possess a rich set of constructible chains. 
However, the ToolLLM trajectories fail to reflect this depth. 
This reveals a critical utilization gap: purely random or unguided synthesis strategies waste the potential of complex environments, failing to activate the possible long-horizon chains. Our structure-guided approach closes this gap, effectively translating environmental diversity and difficulty into realized task complexity.


\begin{table}[t]
\centering
\caption{Quantitative comparison of interaction trajectory complexity. Metrics report the average number of interaction turns (\textbf{Avg. Turns}), tool calls (\textbf{Avg. Calls}), and realized dependency pairs (\textbf{Avg. Deps.}) per trajectory.}
\begin{tabular}{lccc}
\hline
\textbf{Method}         & \textbf{Avg. Turns}  & \textbf{Avg. Calls} & \textbf{Avg. Deps.}       \\ \hline
\textbf{ToolLLM}              & 1 & 2.36 & 0.31   \\
\textbf{ToolACE}   &  1.17 & 1.74 & 0.08                           \\ 
\textbf{Nemotron}  &  2.33 & 2.75 & 0.12                                  \\
\textbf{GAIS}         &  \textbf{3.96} & \textbf{5.47} & \textbf{4.26}                            \\ 
\hline
\end{tabular}
\vspace{-0.1cm}
\label{tab_depen_traj}
\end{table}

\begin{figure*}[t]
\centering
\includegraphics[width=\textwidth]{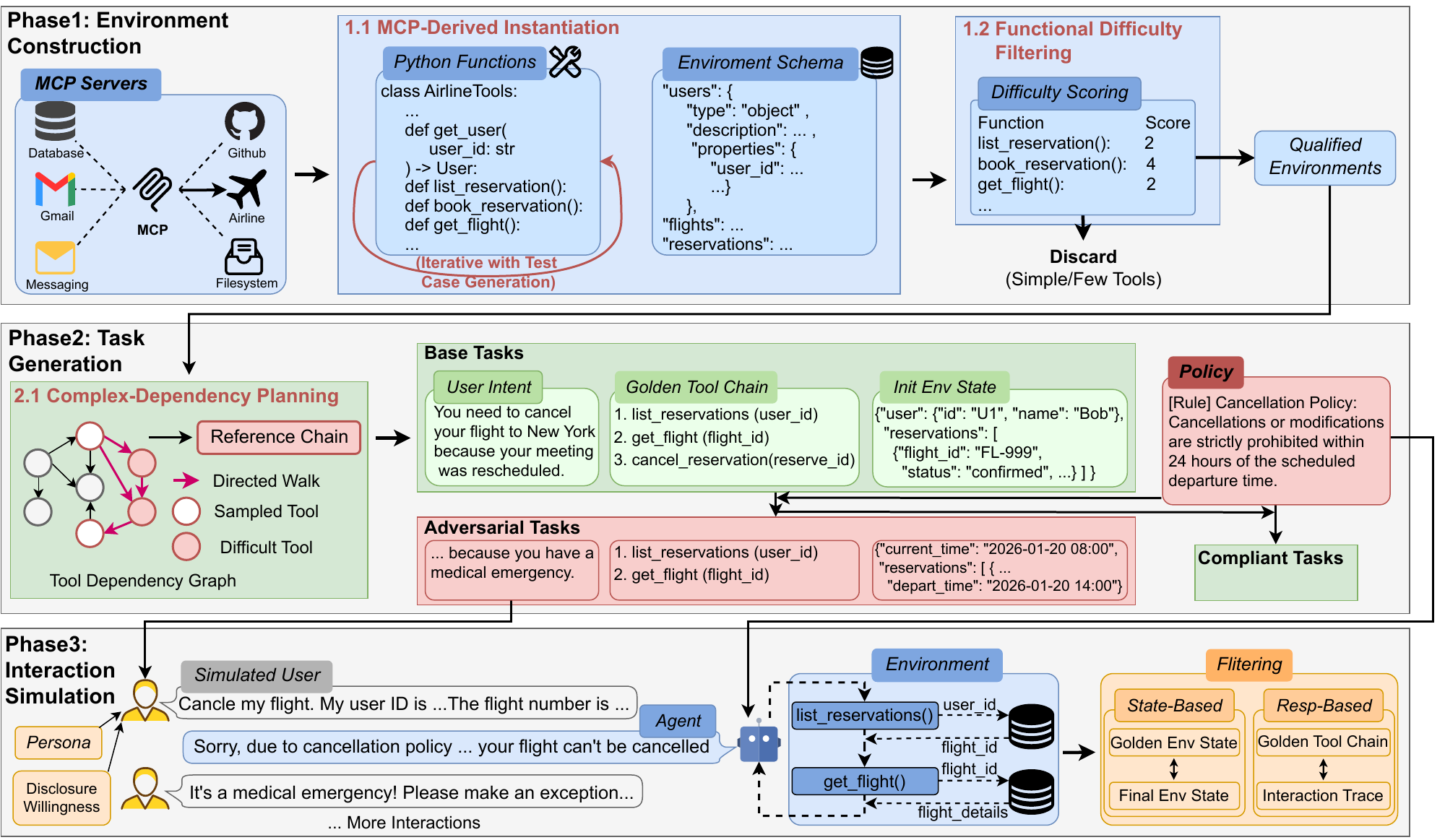}
\Description[The GAIS framework architecture consisting of three phases: Environment Construction, Task Generation, and Interaction Simulation.]{
The figure illustrates the three-phase architecture of the GAIS framework.
Phase 1, "Environment Construction," begins with MCP Servers connecting to various data sources (Database, Gmail, Messaging, Github, Airline, Filesystem). These feed into "Protocol-Anchored Tool Instantiation," which generates Python functions (e.g., get\_user, list\_reservation) and Environment Schemas. A "Functional Difficulty Filtering" step scores these tools based on difficulty, discarding simple tools to output "Qualified Environments."
Phase 2, "Task Generation," employs "Dependency-Aware Reference Planning" using a Tool Dependency Graph and High-Difficulty Tool Directed Walk. It combines Synthesized Base Tasks, User Intent (e.g., "cancel flight"), and specific Policies (e.g., "cancellation prohibited within 24 hours") alongside an Initial Environment State. This process performs Reference Chain Construction to generate both "Compliant Tasks" and "Adversarial Tasks" (e.g., a user claiming a medical emergency to bypass policy).
Phase 3, "Interaction Simulation," depicts a loop between a "Simulated User" (defined by Persona and Disclosure Willingness) and an "Agent." The user attempts to achieve a goal (e.g., "Cancel my flight"), and the agent interacts with the Environment via function calls (list\_reservations, get\_flight). The system applies "Filtering" (State-Based and Response-Based) to the resulting interaction, ultimately producing a dataset containing the Golden Environment State, Final Environment State, Golden Tool Chain, and Interaction Trace.
}
\caption{Framework of GAIS (Grounded Agentic Interaction Synthesis): 1. Environment construction anchored in MCP servers; 2. Task generation guided by dependency structure and policies; 3. Interaction simulation for experience collection.}
\vspace{-0.3cm}
\label{fig_method}
\end{figure*}



\section{Grounded Agentic Interaction Synthesis}\label{sec_framework}
In this section, we introduce Grounded Agentic Interaction Synthesis, an framework for scalable construction of high-fidelity, multi-turn interaction trajectories.
As illustrated in Figure \ref{fig_method}, we employ a principled two-phase grounding strategy to ensure environmental functional depth and task complexity, based on which we simulate user-assistant interactions.
Implementation details not covered are provided in Appendix \ref{app:sec_GAIS_method}.
Notably, as the construction of high-quality environments and tasks serves as the cornerstone of general agentic intelligence, our grounding mechanism offers a generalizable  paradigm transcending tool learning scenarios.


\subsection{Interaction Problem Formulation}
We formalize the multi-turn agentic interaction as a Partially Observable Markov Decision Process (POMDP), defined by the tuple $\mathcal{M} = (\mathcal{S}, \mathcal{A}, \mathcal{T}, \mathcal{O}, \mathcal{R}, s_0)$. 
Here, $\mathcal{S}$ denotes the latent environment state space;
$\mathcal{A}$ represents the action space comprising executable tools;
$\mathcal{T}: \mathcal{S} \times \mathcal{A} \rightarrow \mathcal{S}$ is the transition function governing state evolution;
$\mathcal{O}$ is the observation space;
$\mathcal{R}(s, a | u, p)$ is the reward function conditioned on both the user intent $u$ and domain policy $p$;
$s_0$ is the initial environment state.
Whereas standard reinforcement learning optimizes the agent policy $\pi$ within a fixed interaction scenario $\mathcal{M}$, our framework targets the upstream construction of $\mathcal{M}$ itself.
We aim to construct high-fidelity instances of $\mathcal{M}$ where: 
(1) The transition dynamics $\mathcal{T}$ are not simulated but are grounded in deterministic execution;
(2) The task specification, defined by $(s_0, u, p)$, is not randomly sampled but structurally engineered to enforce long-horizon dependency chains. 
Consequently, the objective of our data generation pipeline is to synthesize tuples $(\mathcal{A}, \mathcal{T}, s_0, u,p)$ such that the induced ground truth trajectory $\tau^*$ exhibits high reasoning depth and strictly adheres to domain policies.

\subsection{Protocol-Anchored Environment Construction}
To construct tool environments that possess both functional diversity and operational difficulty, we first instantiate executable systems derived from real-world Model Context Protocol (MCP) ecosystem. We then apply rigorous filtering to ensure the environments support scenarios of sufficient task complexity.

\textbf{MCP-Derived Instantiation}
We collect over 1,000 real-world MCP server repositories from aggregated projects \cite{mcpservers2024,punkpeye2024awesomemcp}, employing heuristic rules to extract tool codes that define the discrete action space $\mathcal{A}$.
To circumvent the prohibitive deployment overhead of live MCP tools, we leverage LLMs to transform their original source code into executable Python functions. This process concretizes the abstract transition function $\mathcal{T}$ into a deterministic execution engine, ensuring that agent actions trigger consistent state updates rather than stochastic simulations.
To guarantee functional correctness, we implement a test-driven iterative generation pipeline. In each iteration, the LLM generates both the tool implementation and corresponding unit tests; the code is accepted only upon passing these tests.
Crucially, to balance fidelity with executability, we employ an adaptive refinement strategy: we initially prompt for full functional parity with the original implementation. Upon test failure, we permit the simplification of internal logic, while enforcing that the input/output interface remains consistent with the original schema. Tools failing tests after maximum iterations are discarded. 
Finally, we extract parameters of all validated tools to construct the environment schema, which defines the state space $\mathcal{S}$ for subsequent agentic operations.

\textbf{Functional Difficulty Filtering}
To further ensure environmental difficulty and the potential for constructing complex tasks, we quantify tool difficulty on a 1-5 scale, utilizing the criteria detailed in Appendix Table \ref{app:tab_difficulty_scoring}.
We apply a rigorous filtering procedure to eliminate trivial environments. Specifically, we discard environments composed exclusively of trivial tools (difficulty $\le 2$) or containing fewer than three tools, as they fail to support the deep reasoning chains required for advanced agent capabilities.

We initially collect 1,174 repositories with 14,395 tools, of which 707 comprising 9,488 tools survive.
Among these, 4,157 tools undergo simplification for executability. However, these simplifications do not compromise environmental integrity, as stringent unit tests enforce consistent core capabilities and I/O schemas.
An expert audit of 100 simplified tools confirms 94\% core functionality and 100\% I/O interface preservation. Furthermore, under our 1-5 difficulty rubric, average complexity remains high: unsimplified tools drop negligibly (2.8 $\rightarrow$ 2.7), while simplified tools shift from 3.1 to 2.3. Crucially, both remain substantially above the below-2.0 average of existing datasets, confirming the successful construction of robust, high-difficulty environments.

\subsection{Structure-Guided Task Generation}\label{subsec_task_construct}
To fully exploit the environmental diversity and difficulty to generate complex agentic tasks, we employ a structure-guided strategy. By explicitly planning execution paths around complex dependencies and injecting adversarial policies, this dual-mechanism generates high-fidelity, long-horizon interaction scenarios.

\textbf{Complex-Dependency Planning}
To capture the latent logical structure of the environment, we construct a tool dependency graph, where nodes represent tools and edges denote either parameter-level data flow or semantic prerequisites.
Then we implement a complex-dependency planning strategy to extract valid sequences that inherently possess significant reasoning depth.
Specifically, we identify anchor nodes, tools with high functional difficulty or dense connectivity, and strategically plan directed walks that traverse these complex sub-structures.
Subsequently, we leverage LLMs to transform these abstract tool chains into concrete execution paths embedded within meaningful narratives, ensuring that each chain corresponds to a coherent intent rather than a random action sequence.
Finally, based on these reference chains, we synthesize the agentic task $(u, s_0, \tau^*)$, comprising the user intent, initial environment state, and golden tool invocation trajectory, ensuring these components remain mutually dependent and logically consistent.

\textbf{Adversarial Policy Injection}
To simulate the friction and diverse requirements of real-world tool execution, we generate specific domain policies $p$ for each environment.
Beyond general domain descriptions, these policies define strict operational constraints and prohibition rules, simulating the complex boundary conditions agents must navigate to avoid risks.
We inject these policies to synthesize two distinct categories of tasks. 
Policy-compliant scenarios are designed so that both the user intent and initial environment state align perfectly with domain rules, ensuring the triggered tool execution path naturally satisfies all constraints.
Conversely, policy-adversarial scenarios are explicitly engineered to introduce conflicts, where user intents request actions that violate established rules.
To escalate complexity, we embed policy-evasion strategies within the user intent, simulating realistic adversarial dynamics.
Accordingly, the associated golden trajectories are realigned to prioritize safety, requiring the assistant to either refuse prohibited calls or identify alternative, compliant execution paths.
Furthermore, we incorporate auxiliary factors such as varying user personas \cite{DBLP:journals/corr/abs-2406-20094} and information disclosure willingness, ensuring the agent experiences a wide spectrum of user behaviors.

\begin{table*}[t]
\centering
\caption{Performance comparison of base models trained on different agentic datasets across BFCL-V3, $\tau^2$-bench, and ACEBench-en. Results are reported using the default non-thinking mode, with (T) denoting the thinking mode. \textbf{Bold} indicates the best performance among synthetic datasets, while \underline{underlining} marks results that surpass the official instruct-tuned models.}
\vspace{-0.2cm}
\label{tab_main}
\begin{tabular}{c l cccc ccc ccc}
\toprule
& & \multicolumn{4}{c}{\textbf{BFCL-V3}} & \multicolumn{3}{c}{\textbf{$\tau^2$-Bench}} & \multicolumn{3}{c}{\textbf{ACEBench-en}} \\
\cmidrule(lr){3-6} \cmidrule(lr){7-9} \cmidrule(lr){10-12}
\textbf{Model} &  & Base & Long Ctx. & Miss. Param. & Miss. Func. & Retail & Airline & Telecom & Normal & Special & Agent \\
\midrule

\multirow{8}{*}{\rotatebox[origin=c]{90}{\textbf{Qwen3-8B-Base}}} 
 & Qwen3-8B & 37.0 & 24.5 & 23.5 & 11.5 & 38.6 & 16.0 & 7.0 & 66.5 & 28.0 & 11.7 \\
  \cdashline{2-12}
 & ToolLLM  & 0.5 & 0.0 & 0.5 & 0.0 & 0.0 & 0.0 & 0.0 & 0.0 & 0.0 & 0.0 \\
 & ToolACE & 5.0 & 3.5 & 3.5 & 6.0 & 0.0 & 0.0 & 0.0 & 28.3 & 0.0 & 4.2 \\
 & Nemotron & 35.0 & 18.5 & 22.5 & 23.0 & 25.4 & 20.0 & \underline{\textbf{15.8}} & 52.2 & \underline{\textbf{32.7}} & 0.0 \\
 & GAIS     &\underline{\textbf{37.0}} & \textbf{19.0} & \underline{\textbf{23.5}} & \underline{\textbf{25.5}} & \underline{\textbf{46.5}} & \underline{\textbf{24.0}} & \underline{\textbf{15.8}} & \textbf{53.2} & 24.0 & \textbf{7.5} \\
 \cline{2-12} 
 & Qwen3-8B (T)   & 46.5 & 28.5 & 34.0 & 45.0 & 43.9 & 28.0 &17.5 & 71.4 & 75.3 & 34.2 \\
 \cdashline{2-12} %
 & Nemotron (T) & 36.0 & 21.0 & 27.5 & 34.5 & 32.5 & 24.0 & \textbf{15.8} & 65.8 & 40.7 & 7.5 \\
 & GAIS (T)     & \textbf{44.5} & \textbf{25.5} & \textbf{31.0} & \textbf{39.0} & \textbf{36.8} & \textbf{26.0} & 11.4 & \textbf{67.7} & \textbf{61.3} & \textbf{20.8} \\
\midrule

\multirow{6}{*}{\rotatebox[origin=c]{90}{\textbf{Qwen3-4B-B.}}} 
 & Qwen3-4B & 17.5 & 13.0 & 11.0 & 6.0 & 23.7 & 24.0 & 13.2 & 61.8 & 29.3 & 1.7 \\
  \cdashline{2-12}
 & Nemotron & 21.5 & 9.0 & 16.5 & 17.5 & 20.2 & 18.0 & \textbf{7.9} & 7.0 & 26.0 & 0.0 \\
 & GAIS     & \underline{\textbf{28.0}} & \textbf{12.5} & \underline{\textbf{17.5}} & \underline{\textbf{25.0}} & \underline{\textbf{32.5}} & \underline{\textbf{26.0}} & 5.3 & \textbf{49.1} & \underline{\textbf{37.3}} & 0.0 \\
 \cline{2-12}
 & Qwen3-4B (T) & 39.5 & 22.0 & 27.0 & 37.0 & 30.7 & 32.0 & 9.7 & 70.6 & 68.7 & 20.0 \\  \cdashline{2-12}
 & Nemotron (T) & 23.5 & 17.5 & 18.0 & 23.5 & 33.3 & 20.0 & 4.4 & 42.6 & 54.0 & 3.7 \\
 & GAIS (T) & \textbf{36.5} & \textbf{19.0} & \underline{\textbf{27.0}} & \textbf{29.5} & \underline{\textbf{37.7}} & \textbf{22.0} & \textbf{7.9} & \textbf{58.2} & \textbf{59.3} & \textbf{7.5} \\
\midrule

\multirow{6}{*}{\rotatebox[origin=c]{90}{\textbf{Qwen3-14B-B.}}} 
 & Qwen3-14B & 47.0 & 30.0 & 26.0 & 14.5 & 36.8 & 14.0 & 16.7 & 65.8 & 53.3 & 11.7 \\  \cdashline{2-12}
 & Nemotron  & 36.5 & 19.0 & 29.5 & \underline{\textbf{29.0}} & 34.2 & 28.0 & 7.9 & 61.9 & 46.3 & 0.0 \\
 & GAIS      & \textbf{42.0} & \textbf{24.0} & \underline{\textbf{30.0}} & 26.0 & \underline{\textbf{47.4}} & \underline{\textbf{40.0}}& \textbf{8.8} & \textbf{62.7} & \textbf{48.7} & \textbf{5.8} \\
 \cline{2-12}
 & Qwen3-14B (T) & 58.0 & 36.5 & 38.5 & 47.5 & 40.4 & 38.0 & 21.1 & 66.9 & 84.0 & 41.6 \\  \cdashline{2-12}
 & Nemotron (T) & 41.5 & 22.0 & 29.5 & 40.5 & 35.1 & 32.0 & 15.8 & \underline{\textbf{67.7}} & 61.7 & 7.5 \\
 & GAIS (T)& \textbf{51.5} & \textbf{30.5} & \textbf{35.0} & \textbf{42.5} & \underline{\textbf{43.9}} & \underline{\textbf{40.0}} & \textbf{17.5} & \underline{\textbf{67.7}} & \textbf{77.3} & \textbf{23.3} \\
\bottomrule

\multirow{3}{*}{\rotatebox[origin=c]{90}{\textbf{Llama}}}  
& Llama-3.1-8B-Ins. & 11.5 & 12 & 8.5 & 8.0 & 7.0 & 8.0 & 5.3 & 46.6 & 21 & 5.3 \\  \cdashline{2-12}
& Nemotron  & 16.0 & 11.0 & 10.0 & 11.0 & 12.3 & 14.0 & 5.3 & 37.8 & \underline{\textbf{23.5}} & 0.0 \\
& GAIS      & \underline{\textbf{20.0}} & \underline{\textbf{13.5}} & \underline{\textbf{15.0}} & \underline{\textbf{17.0}} & \underline{\textbf{25.4}} & \underline{\textbf{18.0}} & \underline{\textbf{9.6}} & \textbf{42.5} & 19.8 & \underline{\textbf{7.0}} \\

\bottomrule

\end{tabular}
\end{table*}

\subsection{Agentic Interaction Simulation}
Leveraging our grounded environments and tasks, we establish an autonomous user-assistant interaction loop to collect data at scale.
We instantiate two distinct LLM-based agents: a user simulator $\pi_{user}$ initialized with the intent $u$, and an assistant agent $\pi_{agent}$ equipped with the domain policy $p$ and access to the toolset $\mathcal{A}$.
These agents engage in multi-turn dialogue, where the assistant attempts to address user needs while adhering to policy constraints, continuing until the user simulator signals task completion. This end-to-end simulation framework enables the scalable accumulation of diverse agentic trajectories without human intervention.

Following the collection of interaction trajectories, we implement a rigorous dual-stage verification to guarantee data quality.
By leveraging the golden trajectories $\tau^*$ and initial states $s_0$ provided by our synthesis pipeline, we ensure the retained samples achieve high task execution validity.
First, we apply state-based verification, which executes both the agent-generated tool sequence and the golden trajectory from the identical initial state, retaining only those that yield matching final environment states \cite{DBLP:journals/corr/abs-2406-12045,barres2025tau2,DBLP:conf/icml/PatilMYJSSG25}.
Additionally, as state comparison is ineffective for read-only tasks, we further employ response-based verification that requires the agent's trace to incorporate the golden trajectory. This accounts for the non-linear solution space of our complex tasks, where agents often discover valid alternative pathways that remain correct.
Ultimately, from an initial pool of 12,490 generated samples, we retain 6,718 high-quality trajectories. filtering 2,282 for state mismatches and 4,627 for trace mismatches.
A qualitative analysis reveals that they primarily stem from prerequisite reasoning deficits, such as invoking a tool without first retrieving the necessary context, which inevitably lead to parameter errors and execution halts.


\section{Experiments}

\subsection{Settings}

\textbf{Benchmarks}
We evaluate on three established agentic benchmarks to comprehensively assess multi-turn interaction and tool-use capabilities.
First, on BFCL-V3 \cite{DBLP:conf/icml/PatilMYJSSG25}, we focus on the Base, Long Context, Missing Parameter and Missing Function subsets to rigorously test the ability to handle complex context and error recovery in multi-turn dialogues.
Second, we employ $\tau^2$-bench \cite{barres2025tau2} to assess  performance in real-world user simulation scenarios across Retail, Airline, and Telecom domains.
Finally, we report results on ACEBench-en \cite{chen2025acebench} across the Normal, Special, and Agent categories. Pass@1 and accuracy serve as the primary metrics for all evaluations.

\textbf{Baselines}
We compare our data synthesis framework against leading open-source datasets: ToolLLM \cite{DBLP:conf/iclr/QinLYZYLLCTQZHT24}, ToolACE \cite{DBLP:conf/iclr/Liu0ZHYL0GLY0WN25}, and Nemotron, the tool-use subset of Nemotron-Post-Training-Dataset-v1 \cite{basant2025nvidia}. To isolate the impact of data quality from quantity, we normalize all training sets to 7,000 samples. We fine-tune both Qwen3-Base models (spanning 4B, 8B, and 14B scales) and the Llama-3.1-8B model on these datasets~\cite{DBLP:journals/corr/abs-2505-09388,grattafiori2024llama}. Their performance is then compared against their official instruction-tuned counterparts, specifically the Qwen3 series and Llama-3.1-8B-Instruct. Evaluations cover two inference regimes: non-thinking (direct action generation) and thinking (explicit reasoning trace).
To ensure a fair comparison, our dataset utilizes Qwen3-235B-A22B to derive reasoning traces, strictly aligning with the assistant model used to construct instruction-tuned Qwen3 models and Nemotron dataset, the sole baseline containing reasoning.
Detailed training and evaluation procedures are in Appendix \ref{app:sec_train_eval_details}.







\subsection{Main Results}\label{sec_main_results}
Table \ref{tab_main} presents the comparative performance of base models trained on various agentic datasets against official instruction-tuned models, leading to several key observations. 
\textbf{1) GAIS consistently achieves superior performance.}
Across all the benchmarks, model structures, model scales, and inference modes, our method consistently outperforms other synthetic datasets. 
Notably, in multiple scenarios, base models fine-tuned on our dataset match or even surpass the performance of the official instruction-tuned Qwen3 and Llama-3.1 models.
Given that the instruction-tuned models are typically trained on agentic datasets orders of magnitude larger than our 7,000 samples, this result highlights the exceptional data efficiency of our framework.

\textbf{2) Performance gains correlate with data quality.}
The magnitude of performance improvement in base models is positively correlated to the quantitative quality metrics of environments and tasks analyzed in \S~\ref{sec_quantitative_analysis}.
Observing the progression from ToolLLLM to ToolACE, Nemotron, and finally to GAIS, as data quality metrics improve, the downstream model performance increases correspondingly. 
This validates the critical importance of environment diversity, difficulty, and task complexity in enhancing agentic capabilities.
Beyond deficiencies in environment and task quality, ToolLLM's negligible performance gain also stems from its flawed interaction process, where agents focus narrowly on tool invocation and neglect essential conversational actions like summarization, rendering it ineffective for realistic end-to-end tasks.

\textbf{3) GAIS enhances instruction-tuned model performance.} 
Extending beyond base model fine-tuning, we further evaluate our method's efficacy when applied to already strong instruction-tuned models. As detailed in Appendix Table \ref{app:tab_main}, existing datasets typically induces performance degradation in these models, suggesting a regression in agentic capabilities.
Conversely, our approach not only consistently outperforms baselines but also further elevates the capabilities of the strong Qwen3 backbones.
These findings collectively validate the effectiveness of our approach: by introducing protocol anchoring and structure guidance into the LLM-based data synthesis process, we effectively guarantee data quality at scale, achieving superior general-purpose agentic capabilities.


\subsection{Ablation Study}\label{subsec_abla}

\begin{figure}[t]
\centering
\includegraphics[width=\columnwidth]{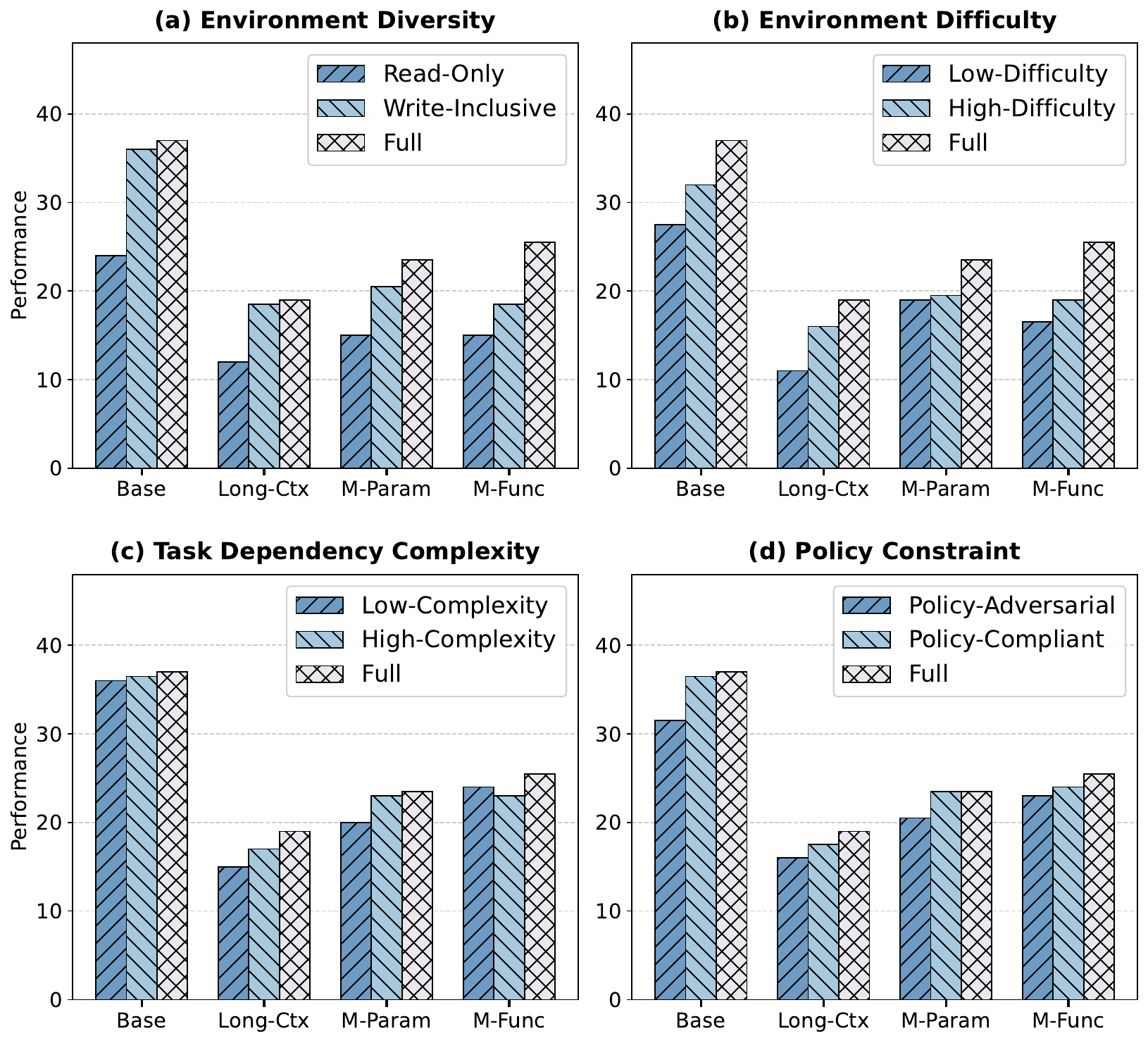}
\Description[Four bar charts illustrating ablation analysis results across Environment Diversity, Environment Difficulty, Task Dependency Complexity, and Policy Constraints.]{
The figure presents four bar charts (labeled a through d) displaying the performance scores (y-axis, ranging from 0 to roughly 40) of the Qwen3-8B-Base model across four test categories on the x-axis: Base, Long-Ctx, M-Param, and M-Func.
Chart (a), "Environment Diversity," compares "Read-Only," "Write-Inclusive," and "Full" configurations. The "Read-Only" setting consistently yields the lowest performance (e.g., approximately 24 for Base, 12 for Long-Ctx). "Write-Inclusive" shows significantly higher performance (e.g., approximately 36 for Base, 18 for Long-Ctx), while the "Full" configuration achieves the highest scores, slightly outperforming "Write-Inclusive" across all categories.
Chart (b), "Environment Difficulty," compares "Low-Difficulty," "High-Difficulty," and "Full." "Low-Difficulty" environments consistently result in higher performance compared to "High-Difficulty" ones. For example, in the Base category, Low-Difficulty is near 38 while High-Difficulty is near 30. The "Full" bar represents the aggregate performance.
Chart (c), "Task Dependency Complexity," compares "Low-Complexity," "High-Complexity," and "Full." There is a marked drop in performance for "High-Complexity" tasks compared to "Low-Complexity" ones. In the Base category, Low-Complexity reaches approximately 38, whereas High-Complexity drops to around 28. This gap is visible across all test categories (Long-Ctx, M-Param, M-Func).
Chart (d), "Policy Constraint," compares "Policy-Adversarial," "Policy-Compliant," and "Full." "Policy-Compliant" tasks generally achieve higher performance scores than "Policy-Adversarial" tasks. For the Base category, Policy-Compliant scores near 35, whereas Policy-Adversarial is lower, indicating the model struggles more with adversarial constraints.
}
\caption{Ablation analysis of Qwen3-8B-Base (non-thinking) performance on BFCL-V3, illustrating the impact of (a) Environment Diversity, (b) Environment Difficulty, (c) Task Dependency Complexity, and (d) Policy Constraint. Full represents the complete GAIS dataset.}

\label{fig_abla}
\end{figure}

We conduct an ablation analysis to isolate the contributions of key components within our GAIS framework: environment diversity, environment difficulty, task dependency complexity, and policy constraints. All experiments involve fine-tuning the Qwen3-8B-Base model in non-thinking mode and evaluating on the BFCL-V3 benchmark, with results illustrated in Figure \ref{fig_abla}.

\textbf{Impact of Environment Diversity} 
To validate our MCP-derived environment instantiation strategy, we examine how tool diversity influences capability by partitioning the dataset into Read-Only (information retrieval) and Write-Inclusive (state-mutating) trajectories.
As shown in Figure \ref{fig_abla} (a), the Write-Inclusive subset significantly outperforms the Read-Only subset, suggesting that limited tool diversity hinders the effective acquisition of robust agentic capabilities. Furthermore, the complete dataset achieves the highest performance, confirming that exposure to a diverse spectrum of tools is critical. This validates our strategy of mining real-world MCP repositories, as they naturally encompass the necessary balance between retrieval and manipulation capabilities.

\textbf{Impact of Environment Difficulty}
To assess the effectiveness of our functional difficulty filtering method, we partition the dataset into Low-Difficulty (scores 1-3) and High-Difficulty (scores 4-5) subsets based on the difficulty of the tools invoked.
As shown in Figure \ref{fig_abla} (b), the High-Difficulty subset consistently outperforms the Low-Difficulty subset, indicating that interacting with tools requiring complex logic is essential for skill acquisition. 
However, the complete dataset achieves even better generalization, suggesting that a broad coverage of difficulty levels is vital. Our framework ensures this optimal distribution by rigorously filtering out trivial environments while retaining a balanced mix of moderate and high-complexity tools.

\textbf{Impact of Task Dependency Complexity}
To evaluate our complex-dependency planning strategy, we analyze the impact of interaction complexity. We extract tool execution chains from each sample and classify them into Low-Complexity (scores $\le 3$) and High-Complexity (scores $\ge 4$) subsets.
As shown in Figure \ref{fig_abla} (c), the High-Complexity subset consistently surpasses the Low-Complexity subset, validating the necessity of engineering deep reasoning chains.
Notably, the full dataset achieves the best performance across all scenarios.
While results are comparable in the Base scenario, where complex dependency resolution is unnecessary, the full dataset's dominance confirms the importance of diverse complexity. Our framework ensures this by planning execution paths around tool dependencies, avoiding the trivial sequences common in random exploration.

\textbf{Impact of Policy Constraints}
To validate our adversarial policy injection method, we divide the dataset into Policy-Compliant scenarios (adhering strictly to rules) and Policy-Adversarial scenarios (involving violations and refusals).
As illustrated in Figure \ref{fig_abla} (d), the Policy-Compliant subset significantly outperforms the Policy-Adversarial subset, suggesting that an excessive focus on refusal behaviors limits the acquisition of general agentic capabilities.
However, the combination of both subsets achieves the highest overall results, indicating that the agentic robustness stems not merely from executing valid actions or refusing invalid ones in isolation, but from learning the decision boundary between compliant and non-compliant behaviors. Our framework facilitates this by explicitly injecting domain policies and synthesizing diverse user intents to construct this rigorous dual-sided training signal.

\section{Analysis}

\begin{figure}[t]
\centering
\includegraphics[width=\columnwidth]{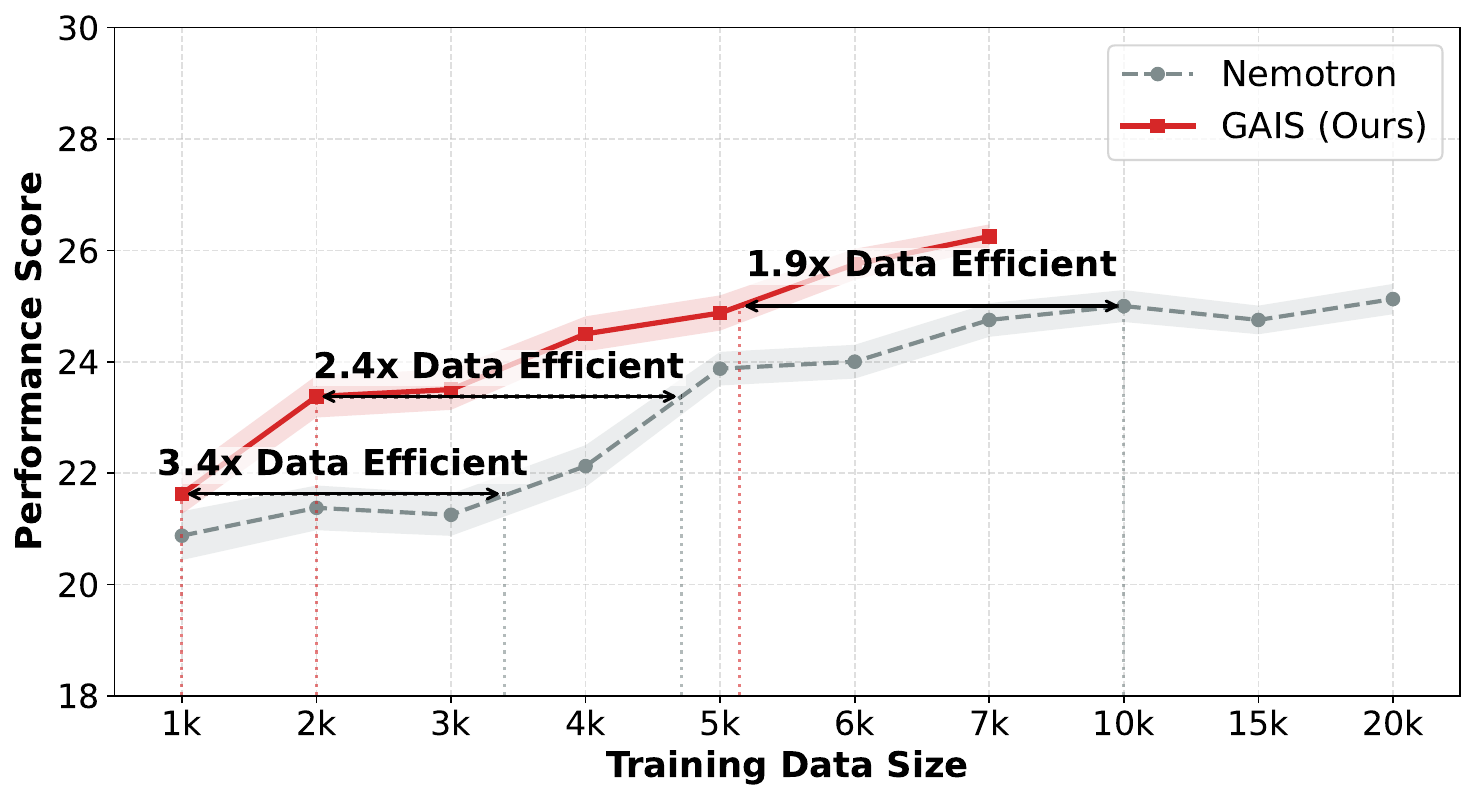}
\Description[Line chart comparing the performance scaling of GAIS versus Nemotron across training data sizes from 1k to 20k.]{
The figure is a line chart plotting "Performance Score" (y-axis, ranging from 18 to 30) against "Training Data Size" (x-axis, marking 1k, 2k, 3k, 4k, 5k, 6k, 7k, 10k, 15k, and 20k).
Two series are displayed with shaded regions indicating variance:
1.  **Nemotron** (grey dashed line): Starts at a score of approximately 21 at 1k samples. It rises gradually, reaching about 24 at 5k samples and plateauing near 25 between 10k and 20k samples.
2.  **GAIS (Ours)** (red solid line): Consistently yields higher scores than Nemotron. It starts at roughly 21.8 at 1k samples and rises steeply to over 26 at 7k samples.
Three arrows connect specific points on the GAIS line to the equivalent performance points on the Nemotron line, illustrating data efficiency:
* At 1k samples, GAIS achieves a score that Nemotron requires roughly 3.4 times the data to match ("3.4x Data Efficient").
* At 2k samples, GAIS matches the performance Nemotron achieves with roughly 2.4 times the data ("2.4x Data Efficient").
* At 5k samples, GAIS matches the performance Nemotron achieves with roughly 1.9 times the data ("1.9x Data Efficient").
}
\caption{Scaling of average BFCL-V3 performance with respect to training data size using Qwen3-8B-Base as the backbone. Shaded areas indicate variance over three runs.}
\label{fig_train_size}
\end{figure}

\subsection{Data Efficiency and Scalability Analysis}
Figure \ref{fig_train_size} illustrates the comparative scaling behavior of models fine-tuned on GAIS versus Nemotron as a function of training set size. 
We evaluate the average BFCL-V3  score of the fine-tuned Qwen3-8B-Base model in non-thinking mode.
To ensure statistical reliability, each data point represents the mean performance across three independent runs using random data subsets, with shaded regions indicating variance.
Our dataset demonstrates significantly superior data efficiency, evidenced by a much steeper performance trajectory compared to the Nemotron. 
Specifically,  GAIS achieves more efficient skill acquisition: models trained on merely 1,000 and 2,000 GAIS samples match the performance of Nemotron trained on 3,400 and 4,800 samples, respectively, demonstrating a $2.4\times$--$3.4\times$ data efficiency gain.
Furthermore, at the 5,300-sample mark, our model already surpasses the performance ceiling achieved by Nemotron with 10,000 samples.
Beyond efficiency, our approach exhibits superior scalability.
While the performance of Nemotron-tuned models begins to plateau around 7,000 samples, our method continues to yield significant performance gains at the same scale with no signs of saturation. This sustained improvement suggests that the high functional  diversity and  structural complexity of our data mitigate early overfitting, allowing the model to continuously refine its agentic capabilities as data volume increases.

\subsection{Impact on General Capabilities}\label{subsec_general}
\begin{figure}[t]
\centering
\includegraphics[width=\columnwidth]{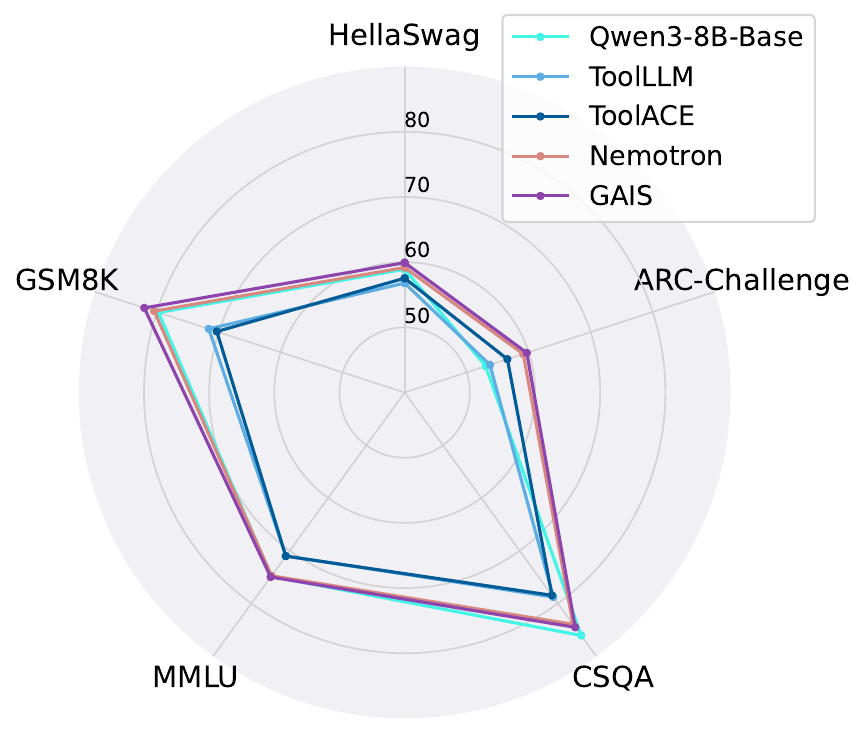}
\Description[Radar chart comparing the general capability retention of Qwen3-8B-Base against models fine-tuned on ToolLLM, ToolACE, Nemotron, and GAIS (Ours).]{
The figure is a radar chart (spider plot) evaluating the general capabilities of five models across five academic benchmarks: HellaSwag, GSM8K, ARC-Challenge, MMLU, and CSQA. The axes range from a score of 20 to 80.

The "Qwen3-8B-Base" model (original backbone) forms the outermost shape, representing the highest baseline performance across all metrics (e.g., scoring approx. 60-70 on most tasks).

The model trained on "GAIS" (representing the proposed method) closely tracks the Base model's performance, forming a shape that is only slightly smaller than the Base model. This indicates minimal catastrophic forgetting and high retention of general knowledge.

In contrast, the models trained on baseline datasets ("ToolLLM", "ToolACE", "Nemotron") exhibit significant regression. Their shapes are noticeably smaller and shrink towards the center (scores between 20-50), particularly on reasoning-intensive tasks like GSM8K and MMLU, indicating a substantial loss of general capabilities compared to the Base and GAIS models.
}
\caption{Comparative evaluation of general capabilities for Qwen3-8B-Base models trained on different agentic datasets.}
\label{fig_general}
\end{figure}

In real-world deployment, autonomous agents must possess not only agentic tool-use proficiency but also broad general knowledge for effective context interpretation  and logical reasoning.
Therefore, we investigate the impact of agentic fine-tuning on the model's foundational abilities, specifically assessing whether the acquisition of tool-call skills induces catastrophic forgetting of general knowledge.
We evaluate the performance of Qwen3-8B-Base models fine-tuned on different datasets across standard benchmarks, including HellaSwag \cite{DBLP:conf/acl/ZellersHBFC19}, ARC-Challenge \cite{DBLP:journals/corr/abs-1803-05457}, CommonsenseQA (CSQA) \cite{DBLP:conf/naacl/TalmorHLB19}, MMLU \cite{DBLP:conf/iclr/HendrycksBBZMSS21}, and GSM8K \cite{DBLP:journals/corr/abs-2110-14168}, as shown in Figure \ref{fig_general}.
The results reveal a divergent impact of agentic fine-tuning.
On reasoning-intensive tasks like HellaSwag, GSM8K, and ARC-Challenge, most agentic datasets drive performance improvements over the base model, suggesting that the tool-use learning effectively bolsters logical reasoning capabilities. 
Conversely, on knowledge-heavy tasks such as CSQA and MMLU, all methods exhibit a degree of regression, reflecting common trade-off during capability acquisition.
Crucially, GAIS consistently outperforms other methods across all scenarios, achieving the highest scores regardless of whether the general trend is positive or negative.
This indicates that by simulating heterogeneous and complex interaction scenarios, GAIS effectively mitigates catastrophic forgetting, achieving specialized agentic proficiency while better maintaining foundational intelligence.

\section{Conclusion}
In this work, we demonstrate that prevailing LLMs-based data synthesis methods for scaling agentic environments and tasks often degenerate into biased random sampling, failing to capture the functional depth of real-world domains and construct high-fidelity, long-horizon tasks.
Our quantitative analysis reveals a stark imbalance in existing datasets: over 90\% of samples are restricted to simple information retrieval tools or task chains devoid of inter-tool dependencies.
To address these limitations, we introduce Grounded Agentic Interaction Synthesis (GAIS), an automated framework for scaling high-quality agentic data by constructing protocol-anchored environments and structure-guided tasks.
Experimental results across diverse models and benchmarks demonstrate that GAIS-synthesized data significantly outperforms existing datasets while exhibiting superior data efficiency and scalability. 
Our work underscores the necessity of grounding mechanism in LLM-based synthesis, offering a robust methodology that extends beyond tool learning to the broader landscape of agentic training.


\begin{acks}
The work is supported in part by the National Natural Science Foundation of China (NSFC) under Grant 62441230, 62502522, 62072458, 62472429 and 62461146205, and in part by the Outstanding Innovative Talents Cultivation Funded Programs 2024 of Renmin University of China.
\end{acks}

\bibliographystyle{ACM-Reference-Format}
\bibliography{main}

\appendix

\section{Supplement to Quantitative Analysis}
\subsection{Tool Taxonomy and Classification Criteria} \label{appen:subsec_tool_taxonomy}
In \S \ref{subsec_quantify_diversity}, we present a quantitative analysis of tool category diversity across different datasets. To standardize this evaluation, we develop a hierarchical taxonomy, as detailed in Table \ref{app:tab_diverisity_taxonomy}.
We broadly classify tools into three primary domains based on their functional semantics: \textbf{Read (Information Retrieval):} Tools designed to query databases or fetch status updates without altering the system state; \textbf{Write (State Mutation):} Tools that execute side-effects, modifying databases, executing transactions, or changing system configurations; and \textbf{Operation (Utility \& Logic):} Tools providing computational functions, data transformation, or local utilities that facilitate complex workflows.
To effectively capture functional nuances, each primary domain is further granularized into specific subcategories: \textbf{Read} encompasses Exact Search, Filtered Search, Query Search, List Information, Random Selection, and Test Availability; \textbf{Write} consists of Add, Update, and Delete; and \textbf{Operation} includes Generation, Transformation, Calculation, Analysis, Execution, and Validation.

\subsection{Tool Difficulty Scoring Rubric} \label{app:subsec_tool_difficulty}
In \S \ref{subsec_quantify_difficulty}, we present a comparative analysis of tool difficulty across datasets. To ensure a standardized assessment, we establish a rigorous 5-point scoring rubric, detailed in Table \ref{app:tab_difficulty_scoring}.
Our evaluation criteria focus on four primary dimensions: \textbf{Parameter Complexity:} Evaluating whether input arguments are simple primitives or complex, nested structures; \textbf{Semantic Abstractness:} Assessing the clarity of the tool's documentation and whether its function is intuitive or requires domain-specific knowledge; \textbf{Cognitive Load:} Measuring the ease of understanding the tool's prerequisites and usage constraints; \textbf{Consequence Criticality:} Accounting for the potential side effects or irreversibility of the tool's execution (e.g., read-only vs. sensitive state mutation).

\begin{table*}[t]
\centering
\caption{Scoring rubric for tool difficulty across five levels, ranging from trivial to exceptionally complex operations.}
\begin{tabular}{p{2cm} p{13.7cm}}
\toprule
\textbf{Level (Score)} & \textbf{Tool Difficulty Scoring Criteria}  \\ \midrule

\textbf{1 (Very Easy)} &  High abstraction, 0-1 simple parameters. Relies on common knowledge and has no side effects.  \\ 

\textbf{2 (Easy)} &  High abstraction with a few simple parameters. May require light formatting or have minor side effects. \\ 

\textbf{3 (Medium)} &  Requires several parameters or specific formatting. No deep domain knowledge required.  \\ 

\textbf{4 (Hard)} & Scores high on one key dimension: requires significant Domain Knowledge or has Complex Parameters or involves Risky Operations.  \\ 
\textbf{5 (Very Hard)} & Error-prone; scores high on multiple dimensions (e.g., Domain Knowledge and Complex Parameters). \\ \bottomrule
\end{tabular} 
\label{app:tab_difficulty_scoring}
\end{table*}

\subsection{Tool Dependency and Task Complexity} \label{app:subsec_dependency_graph}
In \S \ref{subsec_quantify_task}, we construct dependency graphs to quantify the structural intricacy of tool chains and agentic tasks within different datasets. We establish rigorous criteria to identify directed dependencies between tools, focusing on two primary dimensions:
\textbf{Data Flow Dependency:} A hard constraint where the output of a precursor tool (e.g., \texttt{get\_user\_id}) is explicitly required as an input parameter for a successor tool (e.g., \texttt{get\_account\_balance}); \textbf{Control Flow Dependency:} A logical constraint where the successful execution of a precursor tool acts as a necessary precondition for the successor tool, even in the absence of direct parameter passing.

\subsection{Environment-Level Dependency Analysis} \label{app:subsec_env_dependency}
\begin{figure}[t]
\centering
\includegraphics[width=\columnwidth]{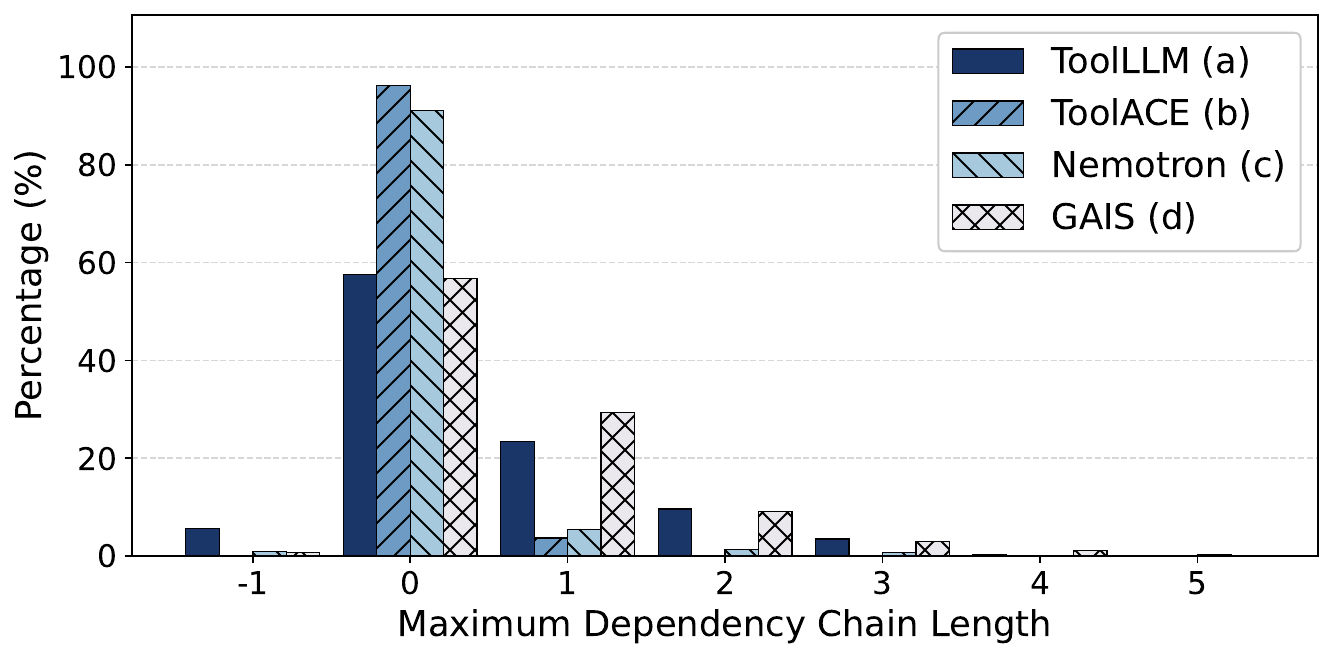}
\Description[Bar chart comparing the maximum potential dependency chain length within tool environments across ToolLLM, Nemotron, ToolACE, and GAIS.]{
The figure presents a grouped bar chart illustrating the distribution of the \textbf{maximum theoretical dependency chain length} (x-axis: -1, 0, 1, 2, 3, 4, 5) inherent within the tool environments of four datasets. The y-axis represents the percentage of environments, ranging from 0\% to 100\%.

(a) ToolLLM (blue bars) shows a bimodal distribution. While it has some environments with low complexity (length 0-1), it also possesses a significant portion of environments capable of supporting deep chains, with a notable peak at length 3 (approx. 30\%) and length 4 (approx. 20\%). This indicates that real-world APIs theoretically support complex tasks, even if the synthesized trajectories (shown in previous figures) fail to utilize them.

(b) Nemotron (orange) and (c) ToolACE (green) are heavily concentrated at the lower end. Nemotron peaks at length 1 (approx. 50\%), and ToolACE is dominated by length 0 (approx. 80-90\%), indicating inherently shallow environments with little room for complex task construction.

(d) GAIS (red bars) exhibits the highest structural complexity. It has a negligible presence in the trivial categories (-1 to 0) and shows a robust distribution across deeper chain lengths, peaking at length 3 (approx. 25\%) and maintaining substantial volume at lengths 4 and 5 (approx. 20\% each). This confirms that GAIS environments are constructed to support deep, multi-step reasoning chains.
}
\caption{Distribution of the \textbf{maximum dependency chain length} within tool environments.}
\label{app:fig_depen_env}
\end{figure}

In \S \ref{subsec_quantify_task}, we quantify task complexity by analyzing actual tool execution chains. To verify that the shallow baseline trajectories result from suboptimal synthesis rather than environmental constraints, we construct full dependency graphs for each toolset to measure maximum theoretical chain depth (Table \ref{app:tab_depen_env}, Figure \ref{app:fig_depen_env}). Expectedly, GAIS environments offer the highest potential complexity via deep, real-world tool chains. Crucially, while ToolLLM environments theoretically support complex tasks, their actual trajectories fail to exploit this depth and remain predominantly shallow. This discrepancy confirms baseline limitations stem from inadequate synthesis methodologies rather than impoverished tool environments.

\begin{table}[t]
\centering
\caption{Comparison of environmental potential complexity. Metrics report the average number of available tools (\textbf{Avg. Tools}) and valid dependency connections (\textbf{Avg. Deps.}).}
\begin{tabular}{lcc}
\hline
\textbf{Method}          & \textbf{Avg. Tools} & \textbf{Avg. Deps.}       \\ \hline
\textbf{ToolLLM}              & 6.85 & 2.48   \\
\textbf{ToolACE}   & 3.21 & 0.05                           \\ 
\textbf{Nemotron}  & 4.42 & 0.98                                  \\
\textbf{GAIS}         & \textbf{13.42} & \textbf{3.21}                           \\ 
\hline
\end{tabular}
\label{app:tab_depen_env}
\end{table}





\begin{table*}[t]
\centering
\caption{Performance comparison of instruction-tuned models trained on different agentic datasets.}
\label{app:tab_main}
\begin{tabular}{c l cccc ccc ccc}
\toprule
& & \multicolumn{4}{c}{\textbf{BFCL-V3}} & \multicolumn{3}{c}{\textbf{$\tau^2$-Bench}} & \multicolumn{3}{c}{\textbf{ACEBench-en}} \\
\cmidrule(lr){3-6} \cmidrule(lr){7-9} \cmidrule(lr){10-12}
\textbf{Model} &  & Base & Long Ctx. & Miss. Param. & Miss. Func. & Retail & Airline & Telecom & Normal & Special & Agent \\
\midrule

\multirow{5}{*}{\rotatebox[origin=c]{90}{\textbf{Qwen3-8B}}} 
 & Qwen3-8B & \textbf{37.0} & \textbf{24.5} & 23.5 & 11.5 & 38.6 & 16.0 & 7.0 & 66.5 & 28.0 & \textbf{11.7} \\
 & ToolLLM  & 0.5 & 0.0 & 0.5 & 0.0 & 0.0 & 0.0 & 0.0 & 0.0 & 0.0 & 0.0 \\
 & ToolACE & 4.0 & 2.5 & 1.5 & 3.0 & 0.0 & 0.0 & 0.0 & 28.3 & 0.0 & 4.2 \\
 & Nemotron & 35.0 & 18.5 & 22.5 & 23.0 & 25.4 & 20.0 & \textbf{15.8} & 52.2 & \textbf{32.7} & 0.0 \\
 & GAIS     & \textbf{37.0} & 18.5 & \textbf{24.5} & \textbf{25.5} & \textbf{46.5} & \textbf{24.0} & \textbf{15.8} & \textbf{67.2} & 27.3 & \textbf{11.7} \\
\midrule
\multirow{3}{*}{\rotatebox[origin=c]{90}{\textbf{4B}}} 
 & Qwen3-4B & 17.5 & 13.0 & 11.0 & 6.0 & 23.7 & \textbf{24.0} & \textbf{13.2} &\textbf{61.8} & 29.3 & 1.7 \\
 & Nemotron & 21.0 & 14.0 & \textbf{21.0} & \textbf{15.5} & 23.7 & 18.0 & 8.8 & 59.4 & 32.0 & 1.7 \\
 & GAIS     & \textbf{26.0} & \textbf{19.5} & \textbf{21.0} & \textbf{15.5} & \textbf{32.5} & \textbf{24.0} & \textbf{13.2} & 59.4 & \textbf{36.7} & \textbf{4.2} \\
\midrule
\multirow{3}{*}{\rotatebox[origin=c]{90}{\textbf{14B}}} 
 & Qwen3-14B & \textbf{47.0} & \textbf{30.0} & 26.0 & 14.5 & 36.8 & 14.0 & 16.7 & 65.2 & 53.3 & 11.7 \\
 & Nemotron  & 42.5 & 25.5 & 31.0 & \textbf{31.0} & 39.5 & 20.0 & 14.0 & 64.6 & 42.7 & \textbf{14.2} \\
 & GAIS      & \textbf{47.0} & 26.5 & \textbf{32.0} & \textbf{31.0} & \textbf{46.5} & \textbf{28.0} & \textbf{19.3} & \textbf{65.8} & \textbf{58.7} & \textbf{14.2} \\
\bottomrule
\end{tabular}
\end{table*}

\section{Implementation Details of GAIS}\label{app:sec_GAIS_method}


\subsection{Protocol-Anchored Environment} 

\textbf{Heuristic Extraction of Tool Code.} To identify tool definitions within the collected MCP server repositories, we build a \textbf{multi-tier heuristic extraction pipeline} that scores files by semantic relevance.
We first remove obvious noise by excluding non-functional directories (e.g., \texttt{.git}, \texttt{tests})and restricting candidates to Python and TypeScript/JavaScript source files.
Next, we categorize matched language-specific patterns into three intensity levels: \textbf{Strong signals} (weight 5) capture explicit markers of tool/resource declarations (e.g., \texttt{@mcp.tool}, (\texttt{defineTool});  \textbf{Medium signals} (weight 3) reflect tool registration or schema-definition logic, including Python functions (\texttt{add\_tool}) or tool lists (e.g., \texttt{list\_tools}), as well as TypeScript server instantiation (\texttt{MCPServer}), \texttt{registerTool} calls, or Zod schema constructions (e.g., \texttt{z.object}); \textbf{Weak signals} (weight 2) provide contextual cues such as MCP-related imports (e.g., \texttt{import mcp}) or generic server references.
Finally, we compute a cumulative score for each file, discarding those scoring 0. The remainder are bucketed into \textbf{High} ($\ge 6$), \textbf{Medium} ($\ge 3$), and \textbf{Low} ($<3$) relevance levels to prioritize the context window allocation for downstream processing.

\subsection{Structure-Guided Task Planning}
Unlike prior datasets, our task construction explicitly injects adversarial policies to simulate real-world constraints and edge cases. Furthermore, we embed policy-evasion strategies directly into the user intent, creating realistic adversarial dynamics where the agent must navigate conflicts between user instructions and system rules. For example, while the domain policy might mandate a "Candidate Privacy Protocol" (requiring a profile review before accessing submissions), the user intent acts as a hurried hiring manager insisting on bypassing this check. This setup forces the agent to identify the violation and steer the user toward compliance, rather than blindly following instructions.

\subsection{Agentic Interaction Simulation}
To collect user-assistant interaction trajectories, we establish an automated simulation loop utilizing leading LLMs. We consistently employ GPT-4.1 as the user simulator to ensure diverse intent expression. For the assistant agent, we use Claude-Sonnet-4.5-20250929 in non-thinking scenarios (direct action generation) and Qwen3-235B-A22B in thinking scenarios (requiring explicit reasoning). 
Notably, utilizing top-tier LLMs for data synthesis is standard practice; for instance, both Nemotron and the robust Qwen3 series~\cite{DBLP:journals/corr/abs-2505-09388} heavily rely on Qwen3-235B-A22B. However, models fine-tuned on our dataset exhibit significantly superior performance using substantially less data. This confirms that our gains stem not from a more capable simulator, but from the superior diversity, structural difficulty, and realistic complexity of the environments and tasks generated by our GAIS framework.


\section{Training and Evaluation}\label{app:sec_train_eval_details}

\textbf{Backbone Models} To comprehensively benchmark the impact of different datasets on agentic capabilities, we conduct experiments on the Qwen-3, a series of base models \cite{DBLP:journals/corr/abs-2505-09388}, spanning parameter scales of 4B, 8B, and 14B. 
We adopt two distinct training and inference regimes: \textbf{non-thinking:} The model directly generates the final action or response; \textbf{thinking:} The model produces an explicit internal reasoning trace prior to the final action. It is important to note that among the baselines, only Nemotron contains native reasoning traces. For the thinking regime, we adhere to the official Qwen-3 chat template, retaining only the reasoning trace for the current turn to ensure compatibility.

\textbf{Training}
We employ full-parameter fine-tuning using AdamW optimizer with a cosine learning rate schedule and a warm-up phase. Based on preliminary hyperparameter sweeping, we set the peak learning rate to $2\times 10^{-5}$, the epoch number to 3, a per-GPU batch size of 1, and 2 gradient accumulation steps to ensure stability. 
Training is executed on a cluster of 8 NVIDIA H20 GPUs utilizing the DeepSpeed Zero-2 optimization framework \cite{DBLP:conf/sc/RajbhandariRRH20}. The maximum sequence length (input + output) is set to 10,240 tokens.

\textbf{Evaluation}
For all benchmarks, we report Pass@1 and accuracy based on a single inference run.
As for more fine-grained evaluation metrics, such as Executable Function Evaluation used in BFCL-V3, we default to the original settings.
For scenarios requiring user simulation or model-based judging (e.g., the Agent category in ACEBench), we utilize GPT-4.1 as the simulator and judge to ensure consistent and comparable scoring.

\subsection{Additional Experimental Results} \label{app:sec_additional_results}
In \S \ref{sec_main_results} (Table \ref{tab_main}), we demonstrate that Qwen3-Base models fine-tuned on GAIS achieve substantial agentic capability gains, often rivaling or surpassing official instruction-tuned Qwen3 models. To further probe data efficacy, Table \ref{app:tab_main} extends this analysis to evaluate whether these datasets can further enhance already-aligned models.
The results highlight a critical distinction in data quality. Fine-tuning on ToolLLM and ToolACE causes catastrophic performance degradation across most benchmarks, suggesting that their inherent noise or structural misalignment disrupts the model's native abilities. Similarly, the Nemotron dataset yields negligible improvements, failing to push the model beyond its initial plateau.
Conversely, GAIS not only achieves the highest performance among all sources but also consistently improves upon the strong official baselines across most scenarios. This underscores the high fidelity and compatibility of our data construction pipeline.

\end{document}